\documentclass{article}
\usepackage[normalem]{ulem}
\useunder{\uline}{\ul}{}
\usepackage{PRIMEarxiv}  
\usepackage[utf8]{inputenc}
\usepackage[table]{xcolor}
\usepackage{enumitem}
\usepackage{array}
\usepackage{url}        

\usepackage{booktabs}   
\usepackage{amsfonts}   
\usepackage{nicefrac}   
\usepackage{longtable}
\usepackage{microtype}  
\usepackage{lipsum}     
\usepackage{graphicx}   
\graphicspath{{media/}} 
\usepackage{caption}    
\usepackage{array}       

\usepackage{multirow}

\usepackage{graphicx}
\usepackage{multicol}    

\usepackage{hyperref}

\title{\textnormal{Fine-Tuning Small Language Models for Domain-Specific AI: An Edge AI Perspective}}

\author{
    \begin{tabular}[t]{c}
    \textbf{Rakshit Aralimatti} \\
    AI Developer \\
    SandLogic Technologies Pvt Ltd \\
    \texttt{rakshit.aralimatti@sandlogic.com} \\[2em]
    \textbf{Kruthika KR} \\
    AI Researcher \\
    SandLogic Technologies Pvt Ltd \\
    \texttt{kruthika.kr@sandlogic.com}
    \end{tabular}
    \hspace{4em}
    \begin{tabular}[t]{c}
    \textbf{Syed Abdul Gaffar Shakhadri} \\
    Lead AI Developer \\
    SandLogic Technologies Pvt Ltd. \\
    \texttt{syed.abdul@sandlogic.com} \\[2em]
    \textbf{Kartik Basavaraj Angadi} \\
    AI Developer \\
    SandLogic Technologies Pvt Ltd \\
    \texttt{kartik.angadi@sandlogic.com}
    \end{tabular}
}

\date{\today}

\begin{document}
\maketitle

\begin{abstract}
Deploying large-scale language models on edge devices faces inherent challenges such as high computational demands, energy consumption, and potential data privacy risks. This paper introduces the Shakti Small Language Models (SLMs)—Shakti-100M, Shakti-250M, and Shakti-500M—which target these constraints head-on. By combining efficient architectures, quantization techniques, and responsible AI principles, the Shakti series enables on-device intelligence for smartphones, smart appliances, IoT systems, and beyond. We provide comprehensive insights into their design philosophy, training pipelines, and benchmark performance on both general tasks (e.g., MMLU, Hellaswag) and specialized domains (healthcare, finance, and legal). Our findings illustrate that compact models, when carefully engineered and fine-tuned, can meet and often exceed expectations in real-world edge-AI scenarios.  
 
\end{abstract}


\section{Introduction}
Recent years have witnessed a surge in large-scale language models (LLMs) such as GPT-3\cite{1} and LLaMA\cite{2}, which exhibit remarkable language comprehension and generation abilities across tasks like summarization, question answering, and creative writing. Despite these capabilities, their deployment typically hinges on substantial computational resources—significant GPU clusters, vast memory requirements, and reliable network connectivity—making them impractical for devices with limited power and storage capacity. Consequently, when applied to settings such as autonomous drones, mobile health services, or on-premise enterprise solutions, the high computational footprint and associated privacy concerns pose serious obstacles. 

A promising direction to alleviate these challenges is Edge AI, wherein models run directly on local hardware, eliminating reliance on remote servers and improving data privacy, latency, and resilience against network failures. However, simply scaling down a massive model often leads to severe compromises in language understanding and output quality. This predicament has prompted the emergence of Small Language Models (SLMs), designed from inception to operate under tight resource constraints while maintaining strong performance. Such models rely on novel architectural techniques (e.g., more memory-efficient attention mechanisms \cite{4}, \cite{5}), quantization to reduce model precision without unduly affecting accuracy [15], and streamlined training or fine-tuning methods\cite{6}, \cite{7}, \cite{8}. 

Within this landscape, the Shakti series stands out for its balanced approach to efficiency and versatility. Building on foundational insights from Shakti-2.5B \cite{shakhadri2024shakti25billionparameter}, the new models—Shakti-100M, Shakti-250M, and Shakti-500M—demonstrate that smaller parameter counts, when combined with robust architectural choices like Rotary Positional Embeddings (RoPE) \cite{3} and specialized attention variants \cite{4}, \cite{5}, can rival or surpass larger models in real-world scenarios. Further, they offer quantized versions (int8, int5, int4) that minimize memory usage and increase tokens-per-second (TPS) throughput, even on devices as constrained as Raspberry Pi boards or entry-level GPUs. 

In parallel with these technical gains, Responsible AI has become paramount. Shakti models incorporate mechanisms to mitigate bias, handle sensitive data privately, and reduce carbon footprints through on-device inference. By pre-training on carefully curated corpora (e.g., Common Crawl \cite{commoncrawl}) and employing fine-tuning strategies such as Supervised Fine-Tuning (SFT) \cite{6}, Direct Preference Optimization (DPO) \cite{8}, and Reinforcement Learning from Human Feedback (RLHF) \cite{7}, these models align not only with user preferences but also with ethical standards in emerging AI regulations. Early evaluations on specialized tasks—such as healthcare QA, finance analytics, and legal contract analysis —suggest that Shakti’s carefully honed parameter sizes and domain-targeted training open up cost-effective, scalable, and privacy-preserving solutions.

\section{Related Work}
\subsection{Large Language Models}

Transformer-based architectures such as GPT-3 \cite{1} and LLaMA \cite{2} brought about a paradigm shift in natural language generation and understanding by massively scaling parameter counts. These models excel in tasks like summarization, translation, and conversational AI. Nonetheless, their reliance on large GPU clusters and extensive memory limits their viability for edge contexts. More recent efforts like Mistral 7B \cite{jiang2023mistral7b} illustrate that parameter-efficient networks can rival larger architectures on certain benchmarks, but they still tend to be too big for tightly constrained devices like smart sensors or mobile ASICs. 

\subsection{Edge and On-Device AI}
Concurrently, Edge AI has emerged as a response to the drawbacks of cloud-dependent ML deployments. Techniques for model compression—such as pruning, knowledge distillation, and quantization—have been widely studied to reduce inference cost. For instance, Big Bird  proposed a sparse attention mechanism to handle long sequences more efficiently, while GQA \cite{5} further refined multi-head attention for memory and computational savings. Although these methods significantly lower resource requirements, most assume the availability of at least moderate GPU capabilities or well-optimized CPU clusters. Truly resource-constrained or battery-powered scenarios necessitate even more streamlined model designs.

\subsection{Small Language Models}
In response to these constraints, Small Language Models (SLMs) adopt an approach that prioritizes efficiency from the outset. This can involve rethinking attention patterns, replacing large embedding layers with more compact structures, and tailoring the training process to smaller parameter regimes. Some SLMs, such as SmolLM \cite{allal2024SmolLM} or Boomer \cite{boomer1b}, \cite{boomer634m}, attempt to compress or distill knowledge from massive teachers. Others, like Shakti-2.5B \cite{shakhadri2024shakti25billionparameter}, reimagine the internal architecture—incorporating, for instance, Rotary Positional Embeddings (RoPE) \cite{3} or advanced attention variants \cite{4}, \cite{5} to retain crucial language capabilities despite fewer trainable parameters. Additionally, techniques like Quantization-Aware Training \cite{chen2024efficientqatefficientquantizationawaretraining} enable these models to operate reliably at int8, int5, or int4 precision, significantly reducing memory usage and power draw.

\subsection{2.4 Responsible AI and Bias Mitigation }
As smaller models expand into consumer devices and sensitive domains, issues of fairness, bias, and toxicity cannot be overlooked. Benchmarks such as BBQ \cite{parrish2022bbqhandbuiltbiasbenchmark} and ToxiGen \cite{hartvigsen2022toxigenlargescalemachinegenerateddataset} help detect unwanted outputs across various demographic and cultural dimensions, and CrowS-Pairs \cite{nangia2020crowspairschallengedatasetmeasuring} measures stereotypical biases in model predictions. Addressing these concerns, SLM developers have begun incorporating Reinforcement Learning from Human Feedback (RLHF) \cite{7} and Direct Preference Optimization (DPO) \cite{8}, alongside data-curation and alignment techniques that target harmful behaviors. By curating training sets and iteratively refining models’ outputs, Shakti and similar SLMs aim to maintain high usability while upholding ethical standards.

\section{Shakti-SLMs Architecture }
The Shakti series—Shakti-100M, Shakti-250M, and Shakti-500M—is designed for efficient, scalable, and adaptable language modeling under edge constraints. Each variant is optimized for different computational budgets: Shakti-100M (10 layers, 640 hidden dim) is suited for ultralightweight applications like IoT and mobile, Shakti-250M (16 layers, 1024 hidden dim) is ideal for domain-specific tasks in finance and healthcare, while Shakti-500M (24 layers, 2048 hidden dim) is tailored for complex multilingual and legal tasks. To optimize memory efficiency, variable grouped query attention (GQA) \cite{5}\cite{shakhadri2024shakti25billionparameter} is used in Shakti-100M and Shakti-250M to reduce key-value projections, whereas Block Sparse Attention \cite{4} in Shakti-500M enables efficient long-context processing. The models integrate Rotary Positional Embeddings (RoPE) \cite{3} for longer sequence modeling without increasing parameter counts and leverage SiLU \cite{10} activation with Pre-Normalization to enhance training stability, particularly for smaller-scale models. For real-time inference, a Sliding Window mechanism \cite{9} inspired by Longformer reuses attention caches to efficiently process long inputs while reducing memory overhead. These design choices ensure that Shakti models maintain strong performance while minimizing computational cost, making them well-suited for deployment in resource-constrained environments. 

\section{Training and Fine-Tuning Methodologies}
The Shakti model series—comprising Shakti-500M, Shakti-250M, and Shakti-100M—undergoes a structured training regimen to optimize performance across various applications. This process includes foundational pre-training, supervised fine-tuning (SFT), and preference alignment through either Reinforcement Learning from Human Feedback (RLHF) or Direct Preference Optimization (DPO). The training methodologies of Shakti-500M, Shakti-250M, and Shakti-100M are built upon the principles outlined in the Shakti-2.5B model\cite{shakhadri2024shakti25billionparameter} but have been adapted to meet the resource constraints and deployment requirements of the smaller variants.

\subsection{Pre-training }
The pre-training process is a foundational phase in training Shakti models, designed to establish a comprehensive understanding of language patterns, grammar, and general knowledge. This phase leverages large-scale, diverse text corpora to expose the models to varied linguistic structures, ensuring adaptability across multiple domains and contexts. Using an unsupervised token prediction approach, the models learn to predict subsequent tokens in sequences, capturing linguistic nuances, contextual relationships, and semantic depth.

Shakti models, built on Transformer-based architectures like their predecessors (e.g., GPT-2 \cite{radford2018gpt} and LLaMA \cite{touvron2023llamaopenefficientfoundation}), utilize the self-attention mechanism to effectively learn dependencies between tokens, even in long sequences. While the general-purpose pre-training corpus includes sources such as Common Crawl and curated datasets, the Shakti-250M model incorporates domain-specific texts to enhance applicability in specialized fields such as healthcare, finance, and legal services. This tailored approach ensures the Shakti-250M model is better equipped to address industry-specific requirements.

A key innovation in the Shakti-500M model is the inclusion of quantization-aware training (QAT) \cite{chen2024efficientqatefficientquantizationawaretraining}. This technique optimizes performance for low-resource devices by reducing memory consumption with low-bit representations while preserving model accuracy, making the model highly efficient for deployment in resource-constrained environments.

By combining large-scale, diverse datasets with targeted domain-specific corpora in Shakti-250M, Shakti models achieve a robust understanding of language, forming a strong foundation for specialized task adaptation. This approach ensures the models generalize effectively across multiple domains while maintaining flexibility for further fine-tuning to address specific use cases.

\subsection{Supervised Fine-Tuning (SFT)}
In the supervised fine-tuning\cite{6} phase, all Shakti models are trained on instruction-specific and task-specific labeled datasets. This process enables the models to adapt their foundational language understanding to specialized applications, ensuring alignment with the unique requirements of domains such as conversational AI, finance, and healthcare. By leveraging labeled datasets, the models refine their outputs, improving their ability to generate accurate, contextually relevant, and domain-specific responses.

\subsection{Reinforcement Learning from Human Feedback (RLHF) }
The Shakti-500M model employs RLHF to fine-tune outputs based on human evaluative feedback, adjusting responses to better align with human preferences regarding relevance, coherence, and accuracy.\cite{7}. This method incorporates feedback from human evaluators to adjust the model's outputs based on criteria such as relevance, coherence, and accuracy. RLHF fine-tunes the model to better align with human preferences, enabling Shakti-500M to deliver responses suitable for complex, multi-turn interactions. This makes it an ideal choice for enterprise-level applications requiring nuanced and human-like conversational abilities.

\subsection{Direct Preference Optimization (DPO)}
In contrast, Shakti-250M and Shakti-100M models employ Direct Preference Optimization\cite{8} (DPO) as a computationally efficient alternative to RLHF\cite{7}. DPO aligns these models with user preferences while minimizing computational resource demands.This approach enables these models to achieve high domain-specific accuracy and deliver quality real-time responses, making them suitable for deployment in resource-constrained environments such as mobile devices and IoT applications. Recent studies have demonstrated that DPO can fine-tune language models to align with human preferences effectively, offering a simpler and more stable alternative to traditional RLHF method. This approach allows Shakti-250M and Shakti-100M to achieve high domain-specific accuracy and deliver high-quality, real-time responses. These attributes make them particularly suited for deployment in resource-constrained environments, such as mobile devices and IoT applications. 

Shakti models have several unique advantages that distinguish them from other models in their category. Deployment flexibility is a key advantage, as Shakti models are uniquely optimized for deployment on low-resource devices, such as smartphones, IoT systems, and wearables. This differentiates them from other larger models, which require significant computational power and are less suitable for edge deployment. Additionally, the inclusion of quantization-aware training during pre-training and fine-tuning makes Shakti models more efficient for int4, int5, and int8 precision deployments, which is not a feature of many other comparative models in the same parameter range. The use of Direct Preference Optimization (DPO) allows Shakti-250M and Shakti-100M to achieve alignment with human preferences at a significantly lower computational cost compared to models relying solely on RLHF. This enables real-time application capabilities without sacrificing quality. 
\section{Dataset Details}

The Shakti-SLM models are trained on extensive and diverse datasets sourced from a wide range of text-rich materials. These datasets provide a broad foundation for language understanding, domain expertise, and adaptability across multiple tasks. The selection and curation of these datasets play a crucial role in ensuring high-quality learning, with detailed dataset information available in the corresponding table.

\begin{itemize}
\item \textbf{Shakti-100M:} The foundational training for Shakti-100M utilizes a large-scale dataset from diverse text-rich sources, providing broad language understanding and domain knowledge. The supervised fine-tuning (SFT) phase refines the model for instructional and conversational tasks, enhancing its ability to follow instructions, generate accurate responses, and assist in specific applications. The DPO stage aligns the model's responses with human preferences, improving output quality and relevance. A detailed list of datasets used for training is available in the corresponding table.

\item \textbf{Shakti-250M:} Shakti-250M is trained on an extensive dataset to establish foundational language knowledge and domain-specific expertise. Pre-training includes general and domain-specific corpora, enabling proficiency in fields like finance and legal applications. The supervised fine-tuning (SFT) phase refines the model for instruction-following and specialized tasks in healthcare, finance, and legal contexts. The DPO stage further fine-tunes outputs to align with preferred behaviors in these domains. A comprehensive list of datasets used in each stage is provided in the corresponding table.

\item \textbf{Shakti-500M:} The Shakti-500M model undergoes pre-training on diverse corpora to develop general language understanding and knowledge across various domains. The supervised fine-tuning (SFT) phase adapts the model for instruction-based applications, enhancing problem-solving, conversational AI, and coding capabilities. RLHF further refines responses through human feedback, ensuring contextual relevance and accuracy. A complete list of datasets used in different training stages is available in the corresponding table.

\end{itemize}

All the datasets used across different training stages for all Shakti models are mentioned in Table \ref{tab:shakti-training-datasets}.

\begin{table}[htbp]
\centering
\small
\renewcommand{\arraystretch}{1.3} 
\begin{tabular}{|p{1.5cm}|p{4.5cm}|p{4.5cm}|p{4.5cm}|}
\hline
 & \multicolumn{1}{c|}{\textbf{Shakti-100M}} & \multicolumn{1}{c|}{\textbf{Shakti-250M}} & \multicolumn{1}{c|}{\textbf{Shakti-500M}} \\
\hline
Pre-Training & 
\begin{itemize}[leftmargin=*]
\item \textit{Common Crawl \cite{commoncrawl}}
\item \textit{Fineweb-EDU-Dedup \cite{skymizer2024fineweb}}
\end{itemize} & 
\begin{itemize}[leftmargin=*]
\item \textit{Fineweb-EDU-Dedup \cite{skymizer2024fineweb}}
\item \textit{AIR-Bench/qa\_finance\_e\_n \cite{airbench2024qafinance}}
\item \textit{Vidhaan/LegalCitationWorthiness \cite{vidhaan2024legalcitation}}
\end{itemize} & 
\begin{itemize}[leftmargin=*]
\item \textit{TxT360 \cite{txt360data2024}}
\item \textit{Common Crawl \cite{commoncrawl}}
\end{itemize} \\
\hline
SFT & 
\begin{itemize}[leftmargin=*]
\item \textit{Cosmopedia v2 \cite{benallal2024cosmopedia}}
\item \textit{Magma-Pro-300K-Filtered-H4 \cite{huggingfacetb2024magpie}}
\item \textit{OpenHermes-2.5-H4 \cite{huggingfacetb2024openhermes}}
\item \textit{Self-oss-instruct-sc2-H4 \cite{huggingfacetb2024starcoder}}
\item \textit{Everyday-conversations-llama3.1-2k \cite{huggingfacetb2024everyday}}
\item \textit{Instruct-data-basics-smolim-H4 \cite{huggingfacetb2024instructdata}}
\end{itemize} & 
\begin{itemize}[leftmargin=*]
\item \textit{lavita/medical-qa-datasets \cite{lavita2024medicalqa}}
\item \textit{ruslannmv/ai-medical-chatbot \cite{ruslanmv2024medicalchatbot}}
\item \textit{axion/pmc\_llama\_instructions \cite{axiong2024pmc}}
\item \textit{windupdate/reddit\_finance\_43\_250k \cite{winddude2024redditfinance}}
\item \textit{Marina-C/question-answer-Subject-Finance-instruct \cite{marinac2024financeqa}}
\item \textit{isacus/open-australian-legal-qa \cite{butler2023openlegalqa}}
\item \textit{mb7419/legal-advice-reddit \cite{mb74192024legaladvice}}
\end{itemize} & 
\begin{itemize}[leftmargin=*]
\item \textit{The Thome \cite{arceeai2024tome}}
\item \textit{Infinity-instruct \cite{baai2024infinityinstruct}}
\end{itemize} \\
\hline
DPO & 
\begin{itemize}[leftmargin=*]
\item \textit{UltraFeedback Binarized \cite{cui2023ultrafeedback}}
\end{itemize} & 
\begin{itemize}[leftmargin=*]
\item \textit{NickyNicky/nano\_finance\_200k\cite{nickynicky2024nano}}
\item \textit{Dhananjay22/legal-dpo \cite{dhananjayg222024dpo}}
\end{itemize} & \\
\hline
RLHF & & & 
\begin{itemize}[leftmargin=*]
\item \textit{UltraFeedback Binarized \cite{cui2023ultrafeedback}}
\end{itemize} \\
\hline
\end{tabular}
\vspace{0.5cm}
\caption{Training datasets used for different Shakti model sizes.}
\label{tab:shakti-training-datasets}
\end{table}


\section{Evaluation and Competitive Study}
\subsection{Comparative Performance Analysis}
In this section, we compare the Shakti series models (Shakti-100M, Shakti-250M, and Shakti-500M) against other leading models in the same or larger parameter range, based on academic benchmark results across a variety of tasks. The comparison provides insights into the relative performance, efficiency, and suitability of the Shakti models for various real-world applications. The performance of the Shakti series models was evaluated on standard benchmarks to ensure consistency and fairness. For comparison models, results from available benchmarks were used, and for those not available, evaluations were conducted by us. 

\subsubsection{Popular Benchmark and Result Analysis for Shakti-100M}

\begin{figure}[h]
\centering
\includegraphics[width=\linewidth]{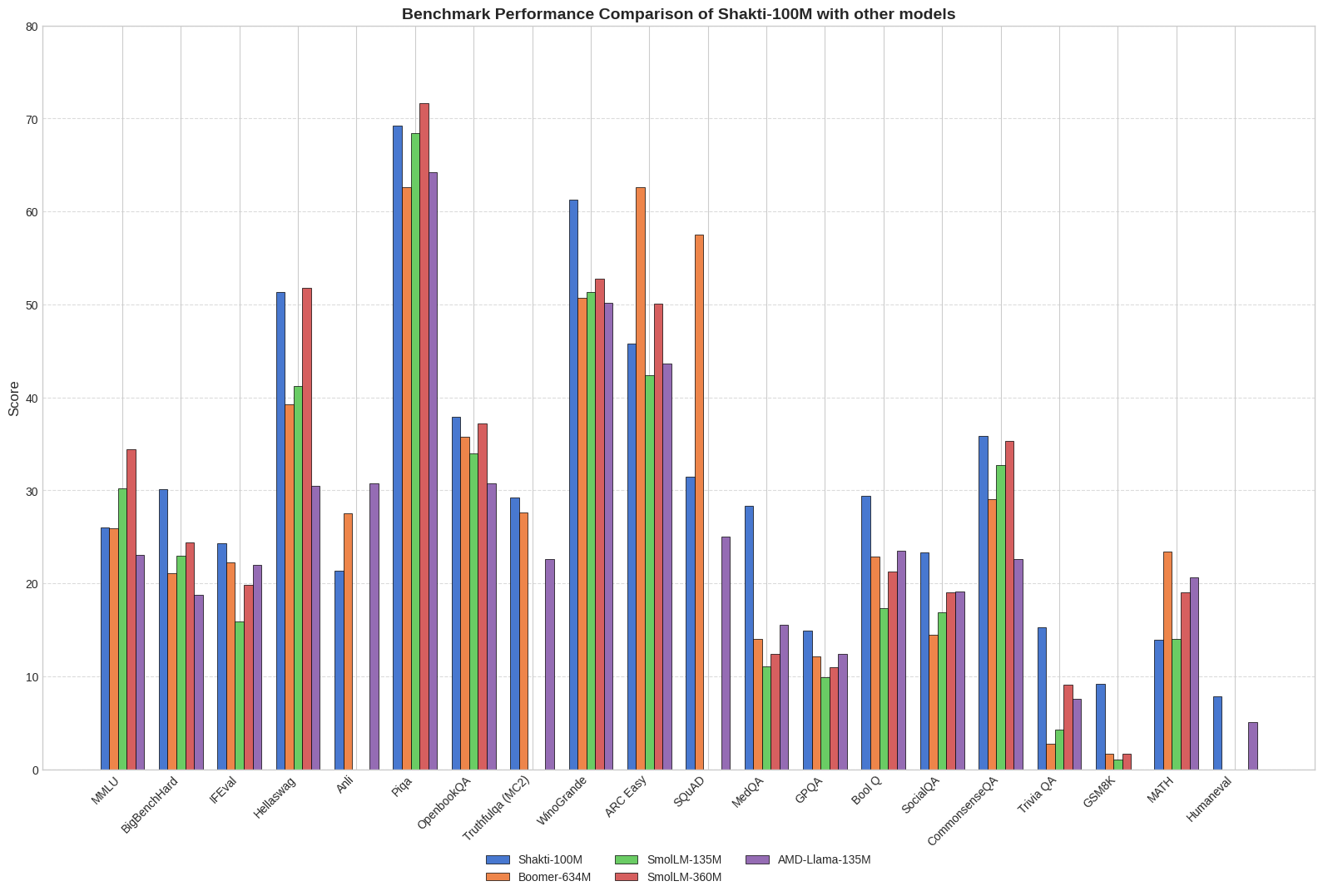}
\caption{Comparison results on academic benchmarks for Shakti-100M, Boomer-634M\cite{boomer634m}, SmolLM-135M\cite{allal2024SmolLM}, SmolLM-360M\cite{allal2024SmolLM}, and AMD-Llama-135M\cite{amdllama135m}, which are in the same parameter range.}
\label{fig:benchmark-100M}
\end{figure}

The Shakti-100M model demonstrates strong benchmark performance, as illustrated in Figure \ref{fig:benchmark-100M},despite being significantly smaller than many competing models. It consistently matches or outperforms larger models in key evaluations, highlighting the effectiveness of its optimized training process on a carefully curated dataset. This approach enables the model to extract intricate patterns and generate accurate predictions, proving that size alone is not the sole determinant of performance. 

A critical factor in Shakti-100M’s success is the balanced size of its pre-training dataset. Models trained on datasets that are either too large or too small often struggle to achieve optimal results. With a 1T token pre-training dataset, Shakti-100M maintains this balance, delivering strong performance across diverse tasks. These results emphasize the importance of strategic data selection and curation in achieving high accuracy and efficiency in language models. 

\subsubsection{Popular Benchmark and Result Analysis for Shakti-250M}
\begin{figure}[h]
\centering
\includegraphics[width=\linewidth]{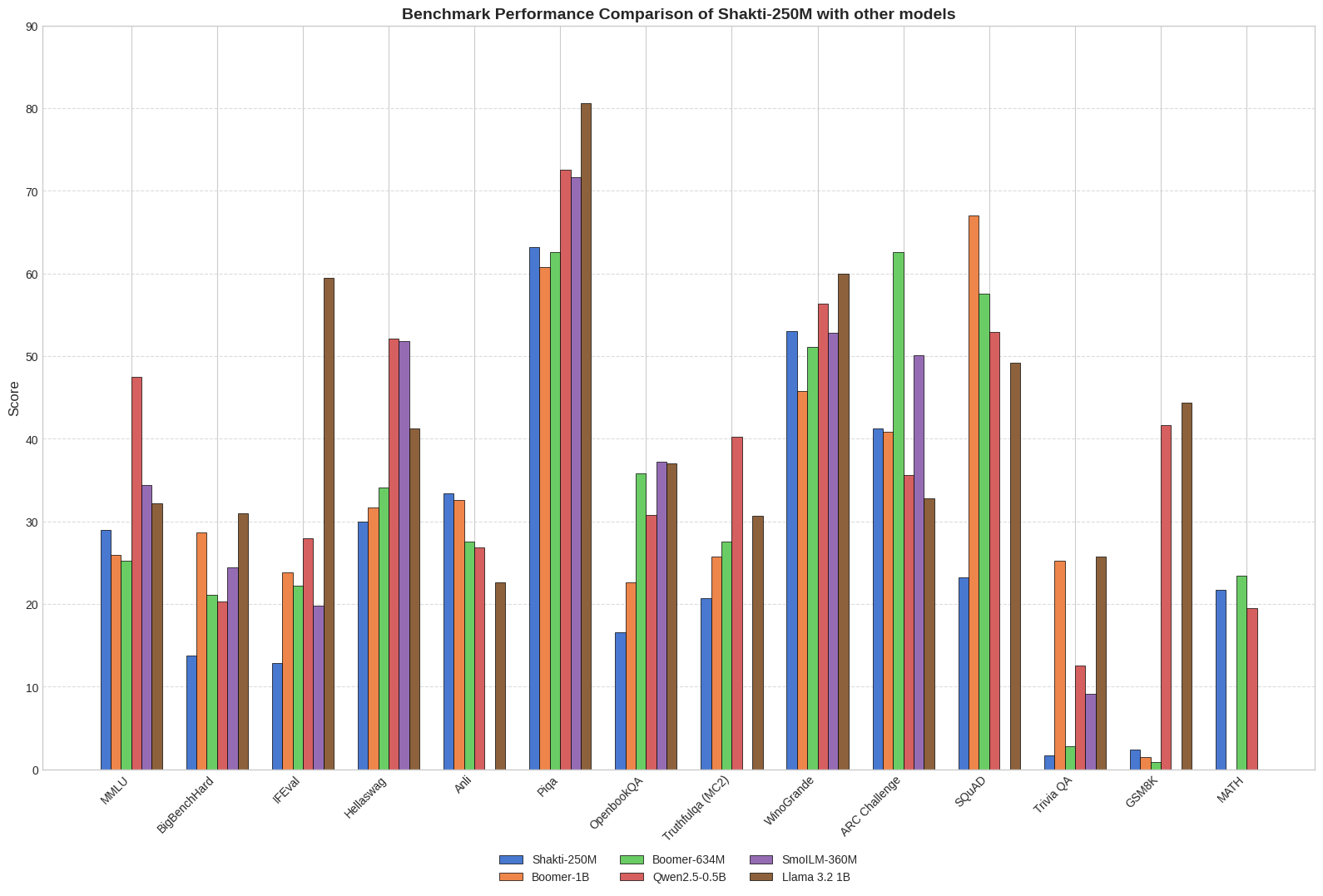}
\caption{Comparison results on academic benchmarks for Shakti-250M, Boomer-1B\cite{boomer1b}, Boomer-634M\cite{boomer634m}, Qwen2.5-0.5B\cite{qwen2.5}, SmolLM-360M\cite{allal2024SmolLM}, and Llama 3.2 1B\cite{llama3.2_1b}. }
\label{fig:benchmark-250M}
\end{figure}

The Shakti-250M model demonstrates outstanding efficiency and performance, competing effectively against larger models, as shown in Figure \ref{fig:benchmark-250M}, such as Boomer-1B \cite{boomer1b} and Llama 3.2 1B \cite{llama3.2_1b}. Despite its smaller size and a more limited training dataset, it achieves impressive results across various NLP tasks. This strong performance highlights its ability to handle diverse language challenges while maintaining computational efficiency.

A key factor behind Shakti-250M’s success is its optimized training process, which leverages clean and well-curated datasets. This approach ensures that the model captures essential linguistic patterns and nuances, enabling high accuracy even with fewer pre-training tokens. While larger models may excel in specific scenarios, Shakti-250M strikes an optimal balance between model size, efficiency, and accuracy.

\subsubsection{Popular Benchmark and Result Analysis for Shakti-500M}
\begin{figure}[h]
\centering
\includegraphics[width=\linewidth]{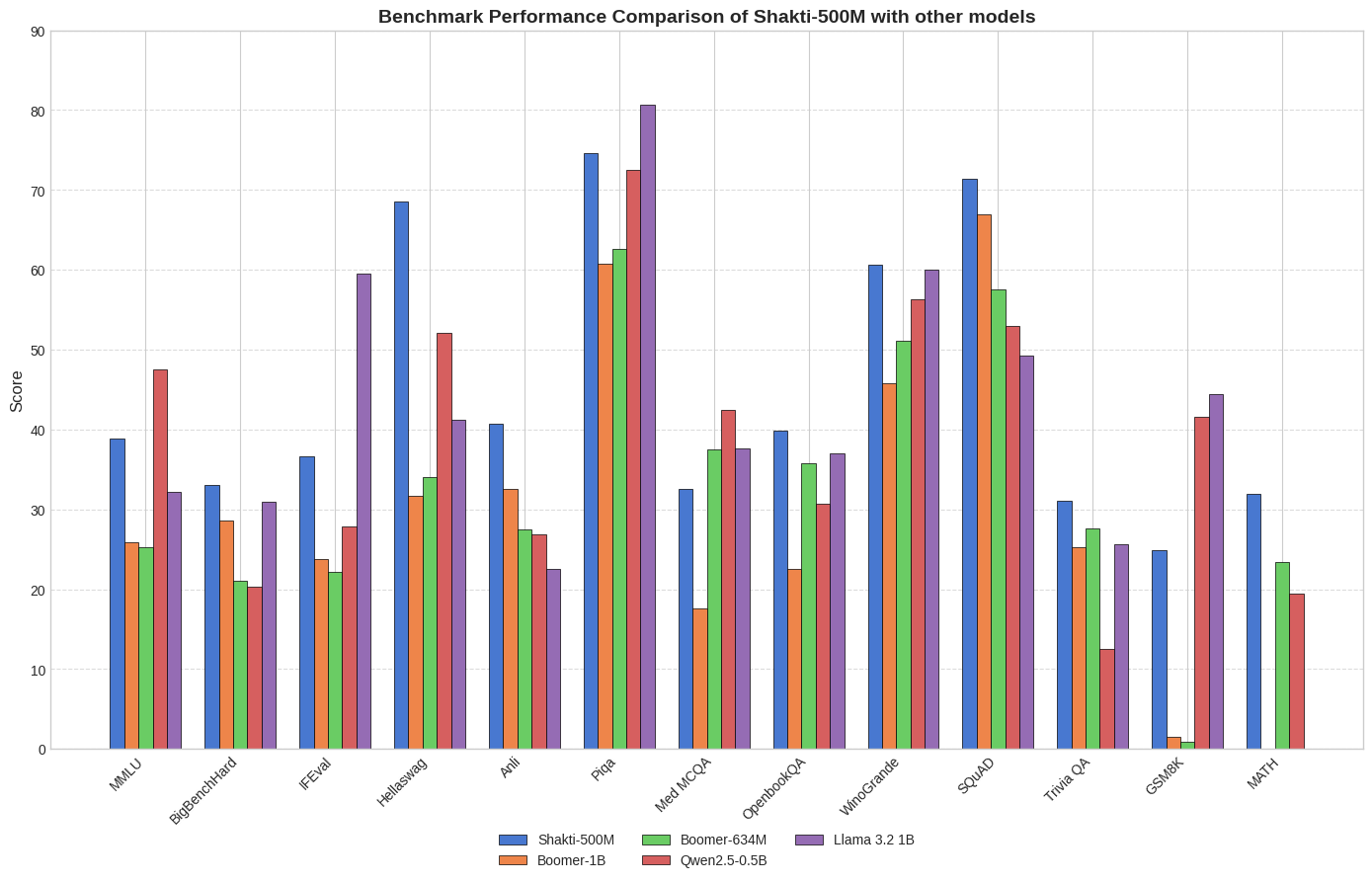}
\caption{Comparison results on academic benchmarks for Shakti-500M, Boomer-1B\cite{boomer1b}, Boomer-634M\cite{boomer634m}, Qwen2.5-0.5B\cite{qwen2.5}, and Llama 3.2 1B\cite{llama3.2_1b}.}
\label{fig:benchmark-500M}
\end{figure}

The Shakti-500M model delivers exceptional performance, as illustrated in Figure \ref{fig:benchmark-500M}, across various NLP tasks, competing effectively with both similar-sized and larger models. Its strong results stem from a well-balanced and carefully curated training dataset, allowing it to maximize the efficiency of its optimized architecture. Despite its relatively smaller size, the model consistently achieves competitive benchmark scores, demonstrating its ability to handle diverse language challenges effectively.

A key contributor to Shakti-500M’s success is its emphasis on data quality and architecture optimization. By leveraging a thoughtfully curated dataset, it achieves high accuracy without relying on excessive model parameters. While larger models may have advantages in specific areas, Shakti-500M maintains an optimal balance between size and efficiency.

\newpage
\subsection{Domain Specific Performance Analysis}

Shakti-250M, tailored with domain-specific training on Finance, Legal, and Healthcare datasets, showcases its specialized capabilities. This section highlights its performance on domain-specific benchmarks and prompt-based evaluation for each domain, emphasizing its efficiency in handling specialized tasks compared to other models.

\subsubsection{Domain Specific Benchmark Result}

\begin{figure}[h]
\centering
\includegraphics[width=\linewidth]{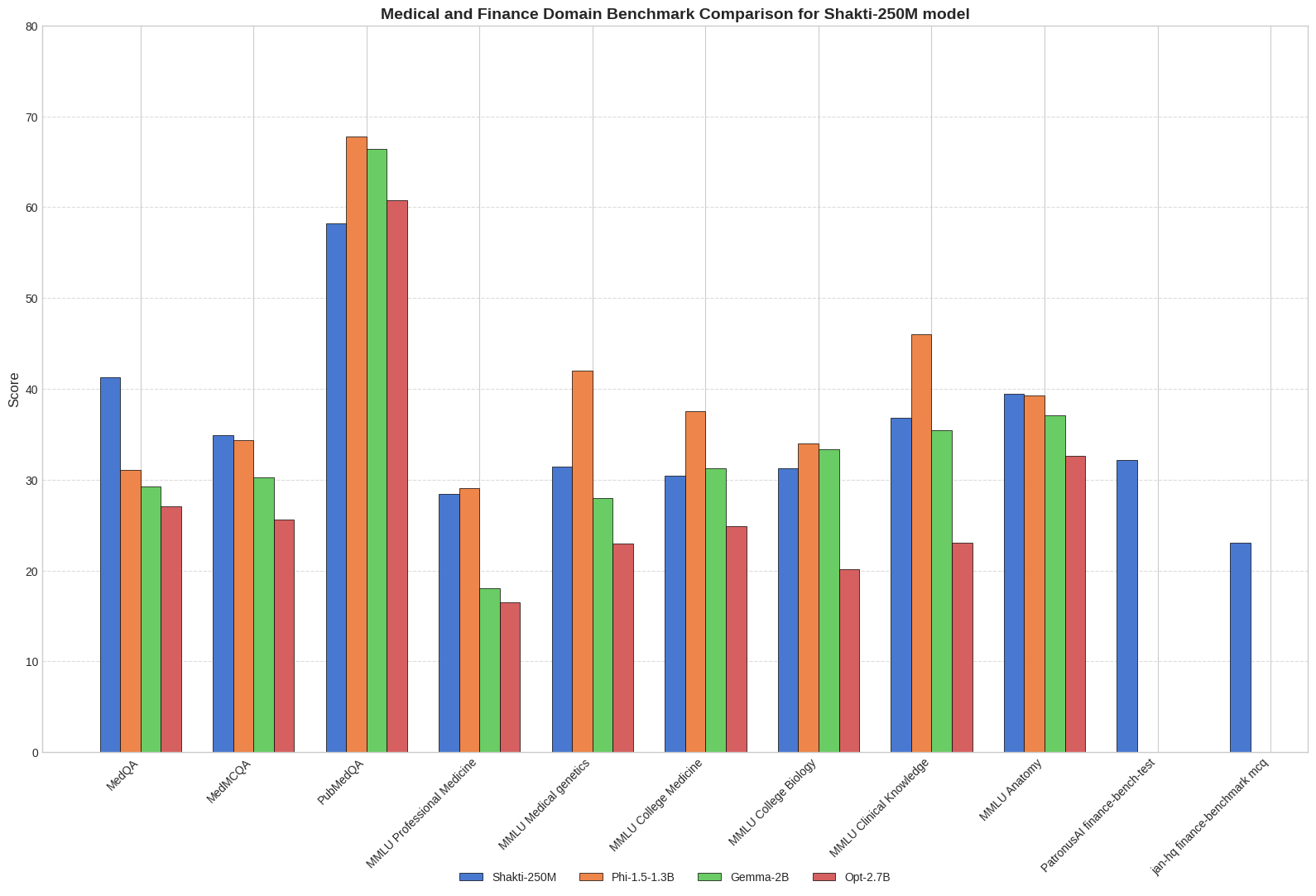}
\caption{Comparison results on medical and finance domain benchmarks for Shakti-250M, Phi-1.5-1.3B\cite{phi15}, Gemma-2B\cite{gemma2b}, and Opt-2.7B\cite{zhang2022opt} models, specifically for the Medical domain.}
\label{fig:medical-finance-benchmark}
\end{figure}

Shakti-250M demonstrates exceptional performance in the healthcare and finance domains, as summarized in Figure \ref{fig:medical-finance-benchmark}, making it a versatile model for domain-specific applications. In healthcare, it excels in tasks requiring complex medical reasoning and shows strong capabilities in understanding and applying clinical knowledge, outperforming expectations of its size.

The model’s compact size and efficiency make it an excellent choice for edge devices and IoT deployment in both healthcare and finance applications. Its ability to deliver reliable and accurate insights under resource-constrained environments opens possibilities for real-time medical assistance, remote diagnostics, on-device health monitoring, financial forecasting, and decision-making tools.

\subsubsection{Prompt-Based Evaluation}  
Table \ref{tab:domain-scores} presents the performance of the Shakti-250M model across healthcare, finance, and legal domains.
\begin{itemize}  
    \item \textbf{Answer Relevancy:} The Shakti-250M model demonstrates strong domain adaptability, achieving Answer Relevancy Scores of 0.85 (healthcare), 0.86 (finance), and 0.81 (legal), showcasing its ability to generate contextually relevant responses.  

    \item \textbf{Summarization Score:} In the legal domain, the model attains a summarization score of 0.86, reflecting its capability to generate concise summaries with moderate fidelity and coverage.  

    \item \textbf{Factual Accuracy:} In the finance domain, Shakti-250M achieves an average factual accuracy score of 0.83, indicating its ability to extract essential information while leaving some room for improvement in precision and detail.  
\end{itemize}

\begin{table}[ht]
    \small
    \centering
    \begin{tabular}{lccc}
        \toprule
        \textbf{Domain} & \textbf{Average Answer Relevancy Score} & \textbf{Summarization Score} & \textbf{Factual Score} \\
        \midrule
        HealthCare & 0.85 & - & - \\
        \midrule
        Legal & 0.81 & 0.86 & - \\
        \midrule
        Finance & 0.86 & - & 0.83 \\
        \bottomrule
    \end{tabular}
    \vspace{0.25cm}
    \caption{Average Answer Relevancy score of Shakti-250M model across domains as mentioned, Summarization score for Shakti-250M model for Legal domain, Average Factual Score for Finance domain.Answer Relevancy: Answer relevancy measures the degree to which the model-generated response aligns with the expected or correct answer, reflecting its accuracy and contextual relevance.Summarization Score: Summarization Score calculates the alignment and coverage of the summary generated for the input.Factual Score: Factual Score evaluates the correctness of factual information in the model's output, measuring how well it captures and reproduces essential details from the input. A score near to 1 indicates better performance of the model in the respective task.}
    \label{tab:domain-scores}
\end{table}

\section{ Shakti-SLMs Multilingual Capabilities}

The Shakti models are designed with robust multilingual capabilities, enabling them to cater to a wide range of linguistic contexts and applications. This is achieved through a specialized tokenizer that supports multiple languages, ensuring efficient representation and processing of diverse linguistic structures. The models can be fine-tuned or aligned with data from various languages, including Indian languages such as Kannada, Hindi, Telugu, and Tamil, as well as widely spoken global languages like Spanish, French, and German. This flexibility makes the Shakti series particularly valuable in multilingual environments, where seamless language adaptation is crucial for effective communication and user engagement. By supporting such a broad linguistic spectrum, the Shakti models democratize access to AI-powered solutions across different regions, breaking language barriers and fostering inclusivity.

\section{Quantization}
Quantization reduces model weight precision from FP32 to lower-bit formats (int4, int5, int8), significantly improving memory efficiency and inference speed while maintaining accuracy. This optimization enables Shakti-100M, Shakti-250M, and Shakti-500M models to run efficiently on resource-constrained hardware, including mobile devices, IoT systems, and drones.

\subsection{Quantization Techniques}
We apply advanced quantization techniques that balance performance and accuracy:
\begin{itemize}
    \item \textbf{Block-wise quantization with scaling factors:} Converting weights into 4-bit (Q4\_0, Q4\_1), 5-bit (Q5\_0, Q5\_1), and 8-bit (Q8\_0) formats.
    \item \textbf{Precision enhancement:} Assigning individual scaling factors to weight blocks.
    \item \textbf{Memory mapping (mmap):} Minimizing RAM usage by directly accessing weights from disk.
    \item \textbf{CPU-specific optimizations:} Accelerating inference using AVX2, ARM NEON, and other architecture-specific instructions.
\end{itemize}

\subsection{Model Size Optimization}
\begin{figure}[htbp]
    \centering
    \includegraphics[width=0.8\textwidth]{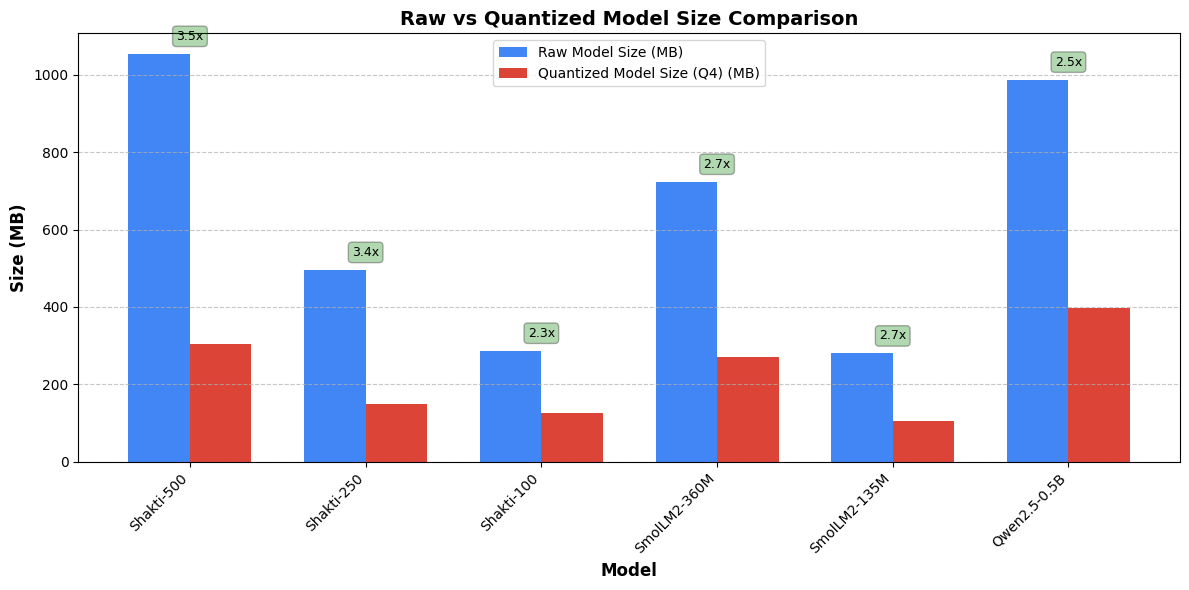}
    \caption{Model size comparison before and after quantization. FP32 represents the original model size, while Q8, Q5, and Q4 represent increasingly aggressive quantization levels. Note the substantial reduction in memory footprint, with Q4 models requiring approximately 8x less memory than their FP32 counterparts.}
    \label{fig:model_size}
\end{figure}

Quantization significantly reduces model size, enabling deployment on resource-constrained devices. Figure \ref{fig:model_size} illustrates the memory footprint reduction across our model variants.

\subsection{Performance Across Hardware Platforms}
\begin{figure}[htbp]
    \centering
    \includegraphics[width=0.9\textwidth]{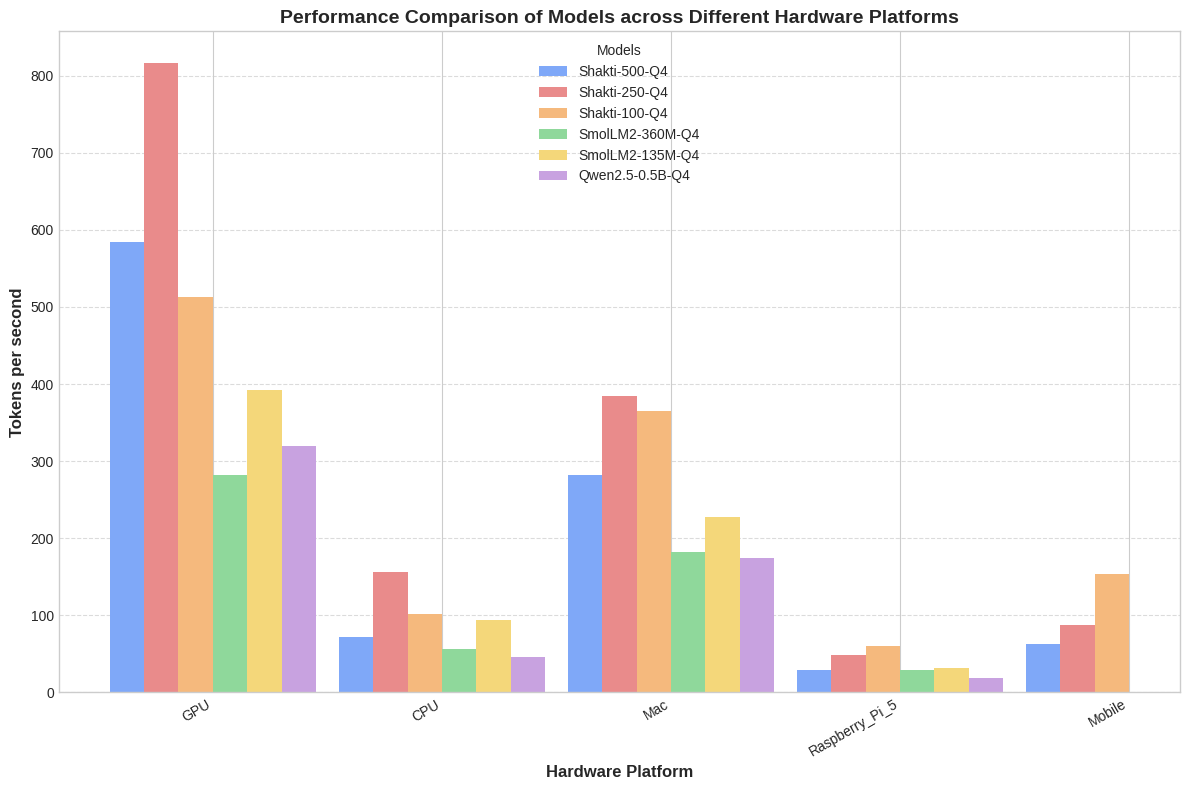}
    \caption{Performance comparison (tokens per second) of Shakti models across different hardware platforms. The graph demonstrates how our models maintain high throughput even on resource-constrained devices compared to similar-sized competitors.}
    \label{fig:performance_comparison}
\end{figure}

Our quantized Shakti models demonstrate strong performance across various hardware platforms, from high-end GPUs to edge devices. Comprehensive testing across multiple device categories reveals superior throughput compared to similar-sized competitor models. Figure \ref{fig:performance_comparison} illustrates the performace of different quantized models across different hardware platform.

\subsubsection{High-Performance Hardware}
\begin{itemize}
    \item \textbf{NVIDIA L40s GPU} (Linux-based VM with AMD EPYC 7R13, 40 GB RAM): Shakti-500-Q4 delivers 583.88 tokens per second (TPS), outperforming SmolLM2-360M-Q4 (281.98 TPS)
    \item \textbf{Intel Xeon Platinum 8488C CPU} (8 cores, 15 GB RAM): Shakti-500-Q4 achieves 72.02 TPS, surpassing Qwen2.5-0.5B-Q4 (45.89 TPS)
    \item \textbf{Apple MacBook Pro} (M3 Max, 36 GB RAM, macOS): Shakti-250-Q4 reaches 385.00 TPS, demonstrating efficiency for general-purpose computing
\end{itemize}

\subsubsection{Resource-Constrained Devices}
\begin{itemize}
    \item \textbf{Raspberry Pi 5} (ARM Cortex-A76, 8 GB RAM, Raspberry Pi OS): Shakti-500-Q4 achieves 29.54 TPS, outperforming SmolLM2-360M-Q4 (28.99 TPS)
    \item \textbf{iPhone 14} (A15 Bionic, 6 GB RAM, iOS 18): Shakti-500-Q4 delivers 62.4 TPS, while Shakti-100-Q4 reaches 153.7 TPS
\end{itemize}

The performance data illustrates that Shakti models are particularly well-suited for deployment in resource-constrained environments while maintaining competitive performance on high-end hardware. This versatility enables real-time AI applications across a wide spectrum of devices, from data centers to edge computing.

\section{ Responsible AI}
The Shakti models embody a strong commitment to Responsible AI principles, addressing key aspects such as fairness, transparency, and environmental sustainability. By leveraging on-device processing, these models prioritize user data privacy, minimizing the risk of exposure to security vulnerabilities inherent in cloud-based systems. The adoption of quantization techniques further reduces the carbon footprint associated with model deployment, aligning with global sustainability goals. Additionally, deliberate efforts to mitigate biases during training enhance the trustworthiness of the models across diverse applications. The Shakti series fosters equitable access to advanced technology through decentralized AI solutions, promoting inclusivity and upholding ethical AI practices.


\subsection{Benchmarking on Responsible AI Datasets}

\begin{itemize}
    \item \textbf{Bias Benchmark for QA (BBQ) \cite{parrish2022bbqhandbuiltbiasbenchmark}:} Shakti-500M achieved 54.08\% accuracy in measuring biases across gender, race, and religion, outperforming Shakti-250M which is 50.2\%.
    
    \item \textbf{ToxiGen \cite{hartvigsen2022toxigenlargescalemachinegenerateddataset}:} On this dataset detecting toxicity across 13 minority groups, Shakti-500M scored 51.5\% accuracy compared to Shakti-250M's 47.5\%, showing improved ability to differentiate toxic content.
    
    \item \textbf{Implicit Hate Speech Dataset \cite{elsherief-etal-2021-latent}:} Shakti-500M reached 69.04\% accuracy in identifying subtle hate speech, significantly higher than Shakti-250M which is 63\%.
    
    \item \textbf{CrowS-Pairs \cite{nangia2020crowspairschallengedatasetmeasuring}:} For bias evaluation across age, disability, gender, and race, Shakti-500M achieved a lower likelihood difference of 3.02 and 51.9\% stereotype percentage versus 3.11 and 52.07\% for Shakti-250M, indicating reduced bias.
\end{itemize}

Results summarized in Tables \ref{tab:ResponsibleAI-benchmark_1} and \ref{tab:ResponsibleAI-benchmark_2} demonstrate Shakti models' capabilities in bias mitigation, toxicity detection, and fairness enhancement.


\begin{table}[ht]
    \small
    \centering
    \begin{tabular}{lccc}
        \toprule
        \textbf{Dataset} & \textbf{Accuracy of Shakti-250M} & \textbf{Accuracy of Shakti-500M}  \\
        \midrule
        BBQ Average & 50.2 & 54.08  \\
        \midrule
        Toxigen & 47.5 & 51.5  \\
        \midrule
        ImplicitHate & 63 & 69.04 \\
        \bottomrule
    \end{tabular}
    \vspace{0.25cm}
    \caption{Accuracy scores of the Shakti models on Bias Benchmark for QA (BBQ), ToxiGen, and Implicit Hate Speech datasets. Higher accuracy indicates the model's improved ability to mitigate biases, detect nuanced toxicity, and accurately classify implicit hate speech, showcasing alignment with Responsible AI principles.}
    \label{tab:ResponsibleAI-benchmark_1}
\end{table}

\begin{table}[ht]
    \small
    \centering
    \begin{tabular}{lccc}
        \toprule
        \textbf{Model} & \textbf{likelyhood diff} & \textbf{pct stereotypes}  \\
        \midrule
        Shakti-250M & 3.11 & 52.07  \\
        \midrule
        Shakti-500M & 3.02 & 51.9  \\
        \bottomrule
    \end{tabular}
    \vspace{0.25cm}
    \caption{Evaluation of the Shakti models on the Crows-Pairs dataset using Likelihood Difference and Percentage of Stereotypes metrics. Lower values in Likelihood Difference indicate reduced preference for stereotypical over non-stereotypical sentences, while a lower Percentage of Stereotypes signifies the model's fairness and ability to minimize bias..}
    \label{tab:ResponsibleAI-benchmark_2}
\end{table}

\section{Conclusion }
The Shakti series of small language models represents a paradigm shift in delivering efficient, secure, and high-performance AI solutions tailored for resource-constrained environments. Designed to address the limitations of large language models, Shakti models—spanning 100M, 250M, and 500M parameters—demonstrate the potential of small language models in enabling real-time, privacy-preserving computation for edge deployment. Leveraging advanced quantization techniques such as int8, int5, and int4, these models minimize memory usage and maximize throughput, achieving exceptional token-per-second performance across diverse hardware platforms, including mobile phones, IoT devices, and GPUs. This makes them ideal for applications requiring low latency and high efficiency, ensuring scalability without compromising accuracy.

Pre-training on diverse, large-scale text corpora equips Shakti models with a deep understanding of linguistic patterns, semantic relationships, and contextual nuances, enabling them to generalize effectively across a wide range of tasks. Fine-tuning techniques, including Supervised Fine-Tuning (SFT) and Direct Preference Optimization (DPO), further refine these models, aligning their outputs with specific domains and tasks. This combination of foundational pre-training and targeted fine-tuning enables the Shakti models to deliver performance comparable to larger language models with significantly fewer parameters, making them ideal for resource-constrained environments. These advantages are consistently reflected in benchmark results, where the models excel in domains such as healthcare, legal, and finance, delivering superior reasoning, factual reliability, and efficiency.

The robust architecture of Shakti models incorporates Rotary Positional Embeddings, Variable Grouped Query Attention, and Block Sparse Attention, ensuring low latency and computational efficiency. These architectural innovations optimize memory usage and enable scalability across a wide range of hardware platforms. The models are also designed to support quantization (int8, int5, and int4), allowing them to achieve high tokens-per-second throughput on devices ranging from mobile phones to GPUs. This quantization-ready design enhances their compatibility with edge devices while maintaining accuracy and performance. Together, the advanced architecture and efficient training methodologies make Shakti models a benchmark for deploying high-performing AI in resource-constrained environments.

Each model in the Shakti series is optimized for specific use cases, showcasing their versatility and adaptability. Shakti-100M, a lightweight, general-purpose model, is tailored for ultra-low-resource devices like smartwatches, consumer electronics, and IoT systems. It excels in tasks such as text summarization, chatbot functionalities, and context-aware assistants, making it indispensable for devices with limited computational resources. Shakti-250M is specifically designed for domain-specific applications in healthcare, legal, and finance sectors, with its ability to operate securely on-premise ensuring data privacy and preventing information leakage. This model is particularly adept at specialized tasks such as patient diagnostics, contract analysis, and financial advising, thanks to its domain-specific fine-tuning. Shakti-500M, the most advanced model in the series, balances general-purpose functionality with enhanced capabilities for complex tasks. With support for multilingual processing and long-context understanding, it is ideal for applications such as customer support chatbots, virtual assistants, and content creation, with deployment potential spanning industries like e-commerce, enterprise communication, and media.

By adhering to Responsible AI principles, Shakti models emphasize fairness, trustworthiness, and accountability. Rigorous data filtering ensures unbiased outputs, while on-device processing enhances privacy and aligns with global sustainability goals by reducing reliance on energy-intensive cloud infrastructures. Use cases across industries highlight the practical impact of Shakti models, from delivering real-time insights in healthcare to powering smart assistants in IoT devices. Their ability to operate securely and efficiently underpins their growing significance in sensitive workflows and privacy-critical environments.

In conclusion, the Shakti series exemplifies the future of edge AI, bridging the gap between state-of-the-art performance and practical deployment. With innovative architecture, efficient quantization, and specialized fine-tuning, these models redefine the capabilities of small language models, making them scalable, privacy-centric, and inclusive. By democratizing access to AI and addressing real-world challenges, the Shakti series sets a new benchmark for sustainable, impactful, and responsible AI deployment across industries.

\section{ Future Scope }

Future developments in the Shakti series aim to enhance multilingual capabilities, particularly in underrepresented languages, to further democratize AI accessibility. Additionally, exploring more efficient training methodologies, such as adaptive pre-training and task-specific finetuning, can further optimize resource consumption. Expanding support for edge computing scenarios, such as integrating Shakti models with federated learning frameworks, could provide robust solutions for collaborative and secure AI deployments. Finally, incorporating advanced feedback mechanisms, such as continuous learning from real-world usage, will improve model alignment and responsiveness in dynamic application settings.


\bibliographystyle{unsrt}  
\bibliography{main}

\clearpage


\end{document}